\documentclass[final]{cvpr}

\usepackage{times}
\usepackage{epsfig}
\usepackage{graphicx}
\usepackage{amsmath}
\usepackage{amssymb}
\usepackage{graphicx}
\usepackage{caption}
\usepackage[nolist]{acronym}
\begin{acronym}[PHOENIX14\textbf{T} ]

\acrodefplural{rnn}[RNNs]{Recurrent Neural Networks}
\acrodefplural{cnn}[CNNs]{Convolutional Neural Networks}
\acrodefplural{hmm}[HMMs]{Hidden Markov Models}
\acrodefplural{gru}[GRUs]{Gated Recurrent Units}
\acrodefplural{crf}[CRFs]{Conditional Random Fields}
\acrodefplural{gan}[GANs]{Generative Adversarial Networks}
\acrodefplural{gpu}[GPUs]{Graphic Processing Units}

\acrodefplural{mdn}[MDNs]{Mixture Density Networks}

\acro{bsl}[BSL]{British Sign Language}
\acro{bleu}[BLEU]{Bilingual Evaluation Understudy}
\acro{blstm}[BLSTM]{Bidirectional Long Short-Term Memory}
\acro{cnn}[CNN]{Convolutional Neural Network}
\acro{crf}[CRF]{Conditional Random Field}
\acro{cslr}[CSLR]{Continuous Sign Language Recognition}
\acro{ctc}[CTC]{Connectionist Temporal Classification}
\acro{dl}[DL]{Deep Learning}
\acro{dgs}[DGS]{German Sign Language - Deutsche Gebärdensprache}
\acro{dsgs}[DSGS]{Swiss German Sign Language - Deutschschweizer Geb\"ardensprache}
\acro{dtw}[DTW]{Dynamic Time Warping}
\acro{fc}[FC]{Fully Connected}
\acro{ff}[FF]{Feed Forward}
\acro{gan}[GAN]{Generative Adversarial Network}
\acro{gpu}[GPU]{Graphics Processing Unit}
\acro{gru}[GRU]{Gated Recurrent Unit}
\acro{gtpt}[G2PT]{Gloss to Pose Transformer}
\acro{hmm}[HMM]{Hidden Markov Model}
\acro{hpe}[HPE]{Hand Pose Enhancer}
\acro{isl}[ISL]{Irish Sign Language}
\acro{lstm}[LSTM]{Long Short-Term Memory}
\acro{mha}[MHA]{Multi-Headed Attention}
\acro{mtc}[MTC]{Monocular Total Capture}
\acro{mse}[MSE]{Mean Squared Error}
\acro{mdn}[MDN]{Mixture Density Network}
\acro{nmt}[NMT]{Neural Machine Translation}
\acro{nlp}[NLP]{Natural Language Processing}
\acro{ph12}[PHOENIX12]{RWTH-PHOENIX-Weather-2012}
\acro{ph14}[PHOENIX14]{RWTH-PHOENIX-Weather-2014}
\acro{ph14t}[PHOENIX14\textbf{T}]{RWTH-PHOENIX-Weather-2014\textbf{T}}
\acro{pof}[POF]{Part Orientation Field}
\acro{paf}[PAF]{Part Affinity Field}
\acro{pttt}[P2TT]{Pose to Text Transformer}
\acro{relu}[RELU]{Rectified Linear Units}
\acro{rnn}[RNN]{Recurrent Neural Network}
\acro{rouge}[ROUGE]{Recall-Oriented Understudy for Gisting Evaluation}
\acro{sgd}[SGD]{Stochastic Gradient Descent}
\acro{sla}[SLA]{Sign Language Assessment}
\acro{slr}[SLR]{Sign Language Recognition}
\acro{slt}[SLT]{Sign Language Translation}
\acro{slp}[SLP]{Sign Language Production}
\acro{smt}[SMT]{Statistical Machine Translation}

\acro{ttgt}[T2GT]{Text to Gloss Transformer}
\acro{ttpt}[T2PT]{Text to Pose Transformer}
\acro{ttp}[T2P]{Text to Pose}
\acro{ttgtp}[T2G2P]{Text to Gloss to Pose}
\acro{wer}[WER]{Word Error Rate}

\end{acronym}
\usepackage{booktabs}
\usepackage[dvipsnames]{xcolor}

\newcommand{\methodName}{\textsc{SignGAN}}

\usepackage{dblfloatfix}
\usepackage[pagebackref=true,breaklinks=true,colorlinks,bookmarks=false]{hyperref}

\def\cvprPaperID{6910} 
\def\confYear{CVPR 2021}

\begin{document}

\title{Everybody Sign Now: \\ Translating Spoken Language to Photo Realistic Sign Language Video}

\author{Ben Saunders, Necati Cihan Camgoz, Richard Bowden\\
University of Surrey\\
{\tt\small \{b.saunders, n.camgoz, r.bowden\}@surrey.ac.uk}
}
\twocolumn[{%
\renewcommand\twocolumn[1][]{#1}%
\maketitle
\begin{center}
    \centering
    \includegraphics[width=.93\textwidth]{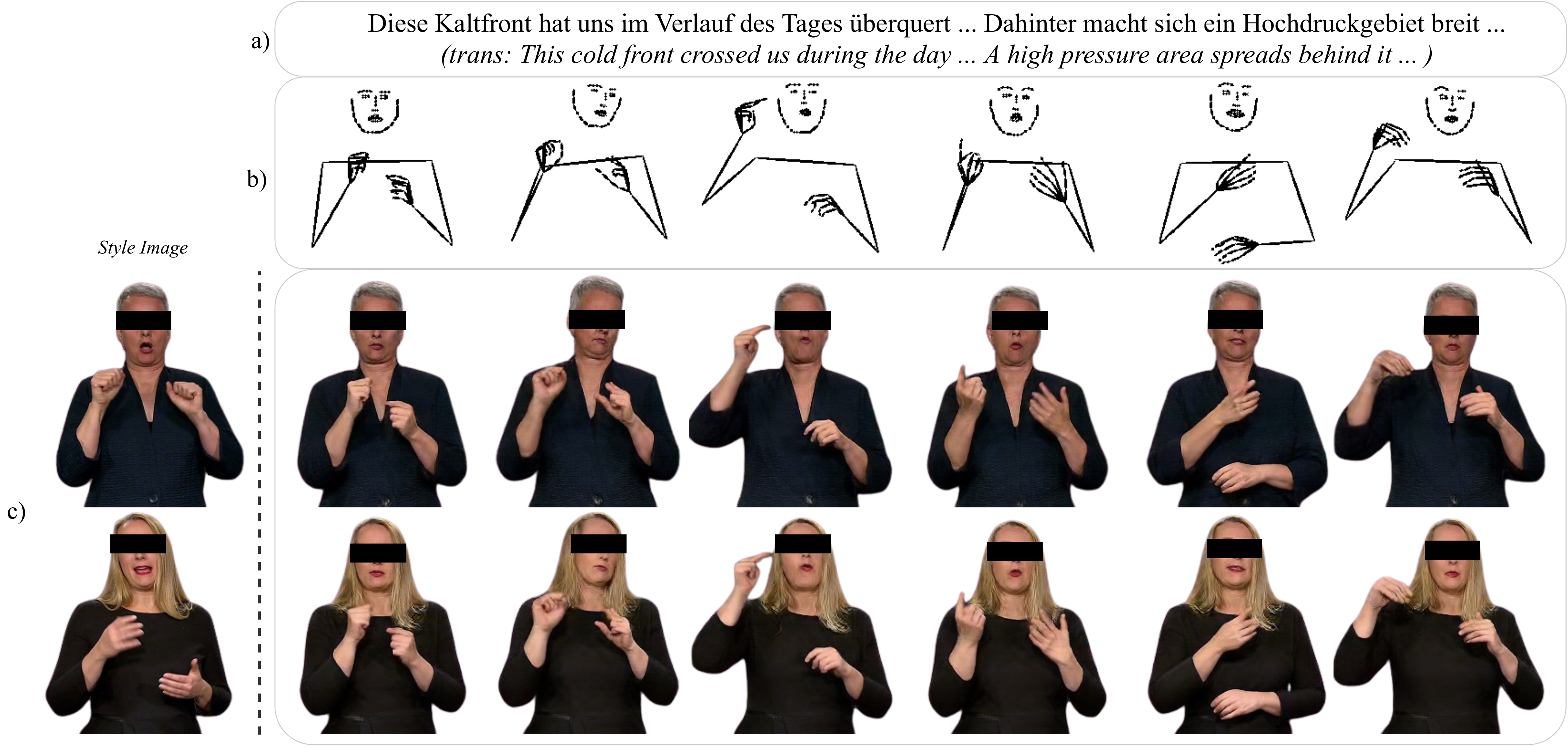}
    \captionof{figure}{\textbf{Photo-Realistic Sign Language Production:} Given a spoken language sentence (a), \methodName{} first produces a skeleton pose sequence (b) and, given a style image, generates a photo-realistic sign language video in the same style (c).}
    \label{fig:intro}
\end{center}%
}]

\begin{abstract}
    To be truly understandable and accepted by Deaf communities, an automatic \acf{slp} system must generate a photo-realistic signer. Prior approaches based on graphical avatars have proven unpopular, whereas recent neural \ac{slp} works that produce skeleton pose sequences have been shown to be not understandable to Deaf viewers.
    
    In this paper, we propose \methodName{}, the first \ac{slp} model to produce photo-realistic continuous sign language videos directly from spoken language. We employ a transformer architecture with a \acf{mdn} formulation to handle the translation from spoken language to skeletal pose. A pose-conditioned human synthesis model is then introduced to generate a photo-realistic sign language video from the skeletal pose sequence. This allows the photo-realistic production of sign videos directly translated from written text.
    
    We further propose a novel keypoint-based loss function, which significantly improves the quality of synthesized hand images, operating in the keypoint space to avoid issues caused by motion blur. In addition, we introduce a method for controllable video generation, enabling training on large, diverse sign language datasets and providing the ability to control the signer appearance at inference.
    
    Using a dataset of eight different sign language interpreters extracted from broadcast footage, we show that \methodName{} significantly outperforms all baseline methods for quantitative metrics and human perceptual studies.
\end{abstract}

\vspace{-1em}

\section{Introduction}

Sign languages are rich visual languages requiring intricate movements of both manual (hands and body) and non-manual (facial) features \cite{stokoe1980sign}. \acf{slp}, the translation from spoken language sentences to sign language sequences, must replicate these intricate movements to be truly understandable by Deaf communities. Prior deep-learning approaches to \ac{slp} have mainly produced skeleton pose sequences \cite{saunders2020adversarial,saunders2020progressive,zelinka2020neural}, whereas Deaf viewers have been shown to prefer synthesized sign language videos \cite{ventura2020can}. 

Even though the field of pose-conditioned human synthesis has progressed significantly \cite{balakrishnan2018synthesizing,chan2019everybody,wang2019few,yang2018pose}, the application of techniques such as \textit{Everybody Dance Now} \cite{chan2019everybody} perform poorly in the \ac{slp} task \cite{ventura2020can}. Specifically, generated hands are of low quality, leading to a sign language production that is not understandable by the Deaf. 

\begin{figure*}[t!]
    \centering
    \includegraphics[width=1.00\linewidth]{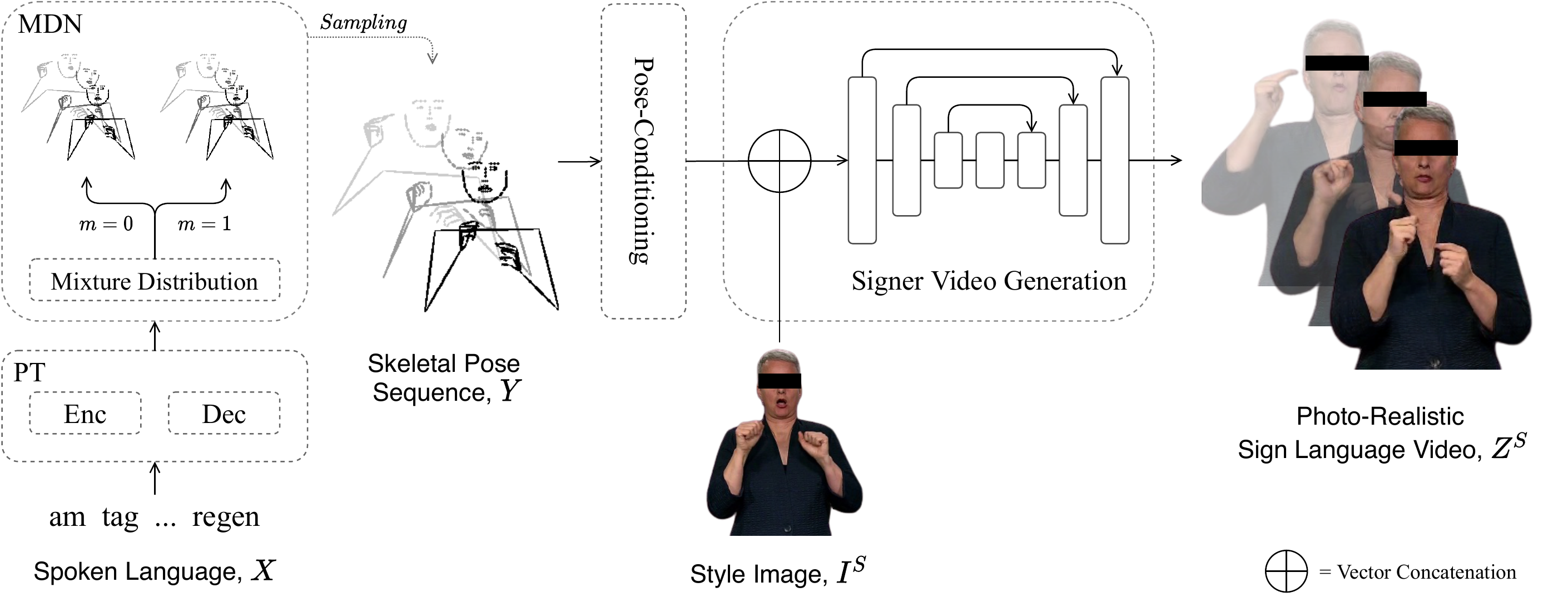}
    \caption{\methodName{} network overview, showing the multi-stage generation of a photo-realistic sign language video, $Z^{S}$, given a spoken language sentence, $X$, and style image, $I^{S}$. (PT: Progressive Transformer, MDN: Mixture Density Network)}
    \label{fig:model_overview}
\end{figure*}

In this paper, we propose \methodName{}, the first \ac{slp} model to produce photo-realistic continuous sign language videos directly from spoken language. Firstly, we translate from spoken language sentences to continuous skeleton pose sequences using a transformer architecture. Due to the multi-modal nature of sign languages, we use a \acf{mdn} formulation that produces realistic and expressive sign sequences (Left of Figure \ref{fig:model_overview}).

The skeleton pose sequences are subsequently used to condition a video-to-video synthesis model capable of generating photo-realistic sign language videos (Right of Figure \ref{fig:model_overview}). Our network is the first to jointly solve the challenging tasks of continuous sign language translation and photo-realistic generation in a single neural pipeline. 

Due to the high presence of motion blur in sign language datasets \cite{forster2014extensions}, a classical application of a discriminator over the hands could potentially lead to an increase in blurred hand generation. To avoid this, we propose a novel keypoint-based loss that uses a set of \textit{good} hand samples to significantly improve the quality of hand image synthesis in our photo-realistic signer generation module. 

To enable training on diverse sign language datasets and to make full use of the variability in appearance across signers, we propose a method for controllable video generation. This allows \methodName{} to model a multi-modal distribution of sign language videos in different styles, as shown in Figure \ref{fig:intro}, which has been highlighted as important by Deaf focus groups \cite{kipp2011assessing}. In addition to providing a choice to the user about the signer's appearance, this approach provides improved definition of the hands and face (as discussed later).

We evaluate on the challenging \ac{ph14t} corpus for continuous \ac{slp}, achieving state-of-the-art back translation results. Furthermore, we collect a dataset of sign language interpreters from broadcast footage, for photo-realistic generation. We compare against state-of-the-art pose-conditioned synthesis methods \cite{chan2019everybody,stoll2020text2sign,wang2018video,wang2018high} and show that \methodName{} outperforms these approaches in the signer generation task, for quantitative evaluation and human perception studies.

The rest of this paper is organised as follows: In Section~\ref{sec:related_work}, we review the previous literature in \ac{slp} and human synthesis. In Section~\ref{sec:methodology}, we outline the proposed \methodName{} network. Section~\ref{sec:experiments} presents quantitative and qualitative model comparison, whilst Section \ref{sec:conc} concludes.

\section{Related Work} \label{sec:related_work}

\paragraph{Sign Language Production}

Sign language has been a focus of computer vision researchers for over 30 years \cite{bauer2000video,starner1997real,tamura1988recognition}. However, research has primarily focused on initially isolated sign recognition \cite{antonakos2012unsupervised,grobel1997isolated,ozdemir2016isolated}, increasingly \ac{cslr} \cite{camgoz2017subunets,cui2017recurrent,koller2015continuous,koller2019weakly} and, only recently, the task of \ac{slt} \cite{camgoz2018neural,camgoz2020multi,camgoz2020sign,ko2019neural,orbay2020neural,yin2020sign}.

\acf{slp}, the translation from spoken to sign language, has been historically tackled using animated avatars \cite{cox2002tessa,karpouzis2007educational,mcdonald2016automated}. However, there has been an increase in deep learning approaches to \ac{slp} \cite{saunders2020adversarial,saunders2020progressive,stoll2018sign,xiao2020skeleton,zelinka2020neural}. Initial attempts have focused on the production of concatenated isolated signs that disregard the grammatical syntax of sign language \cite{stoll2020text2sign,zelinka2020neural}. Saunders \etal proposed the first \ac{slp} model to produce continuous sign language sequences direct from source spoken language \cite{saunders2020progressive}. A \textit{Progressive Transformer} model was introduced that uses a counter decoding technique to predict continuous sequences of varying length. We expand this model with a \ac{mdn} formulation for more expressive sign production.

However, these works represent sign language as skeleton pose sequences, which have been shown to reduce the Deaf comprehension compared to a photo-realistic production \cite{ventura2020can}. Stoll \etal produced photo-realistic signers, but using low-resolution, isolated signs that do not generalise to the continuous domain \cite{stoll2020text2sign}. In this work, we produce high-resolution photo-realistic continuous sign language videos directly from spoken language input. To improve the quality of production, we further enhance hand synthesis and introduce the ability to generate multiple signer appearances.

\paragraph{Pose-Conditioned Human Synthesis}

\acfp{gan} \cite{goodfellow2014generative} have achieved impressive results in image \cite{isola2017image,radford2015unsupervised,wang2018high,zhu2017unpaired} and, more recently, video generation tasks \cite{mallya2020world,tulyakov2018mocogan,vondrick2016generating,wang2019few,wang2018video}. Specific to pose-conditioned human synthesis, there has been concurrent research focusing on the generation of whole body \cite{balakrishnan2018synthesizing,ma2017pose,siarohin2018deformable,tang2020xinggan,zhu2019progressive}, face \cite{deng2020disentangled,kowalski2020config,zakharov2020fast,zakharov2019few} and hand \cite{liu2019gesture,tang2018gesturegan,wu2020mm} images. 

However, there has been no research into accurate hand generation in the context of full body synthesis, with current methods failing to generate high-quality hand images \cite{ventura2020can}. Due to the hands being high fidelity items, they are often overlooked in model optimisation. In addition, most prior work has been conditioned on poses that do not contain detailed hand poses, leading to the generation of blurry hand images \cite{balakrishnan2018synthesizing,wang2018video}. Chan \etal introduced FaceGAN for high resolution face generation \cite{chan2019everybody}, but no similar work has been proposed for the more challenging task of hand synthesis. However, simply adding a \ac{gan} loss on the hands serves no purpose if the original data is also blurred. In this work, we propose a keypoint-based loss using a set of \textit{good} hand samples to enhance hand synthesis.

The task of human motion transfer, transferring motion from source to target videos via keypoint extraction, is relevant to our task \cite{chan2019everybody,wei2020gac,zhou2019dance}. However, there has been limited prior research into the generation of novel poses not seen in source videos or conditioned on a given input. Previous work has been restricted to conditioning pose generation on a given action \cite{yang2018pose} or audio \cite{ferreira2020learning,ren2020self}. Our model first generates a sequence of unseen sign language poses given a spoken language sentence, which are subsequently used to condition our human synthesis module.

The ability to generate multiple styles at inference by separately controlling appearance and content is an important aspect of realistic human synthesis. Recent works have used dynamic weights to produce unseen appearances in a few-shot manner \cite{wang2019few,zakharov2019few}, but continue to produce only a single style at inference. A semantic consistency loss for example-driven generation \cite{wang2019example} has also been proposed, but this approach requires a manual labelling of style-consistency in the dataset and increases network computation. In this work, we introduce a novel method for controllable video generation, to enable training on large, diverse sign language datasets and support the synthesis of multiple styles from a single model.

\section{Methodology} \label{sec:methodology}

Given a source spoken language sentence, \hbox{$X = (x_{1},...,x_{U})$} with $U$ words, and a style \hbox{image, $I^{S}$,} our goal is to produce a photo-realistic sign language translation video in the same style; $Z^{S} = (z^{S}_{1},...,z^{S}_{T})$ with $T$ time steps. We approach this problem as a multi-stage sequence-to-sequence task and propose the \methodName{} network. Firstly, the spoken language sentence is translated to a sequence of sign language poses, \hbox{$Y = (y_{1},...,y_{T})$}, as an intermediate representation. Next, given $Y$ and $I^{S}$, our video-to-video signer generation module generates a photo-realistic sign language video, $Z^{S}$. Figure \ref{fig:model_overview} provides an overview of our network. In the remainder of this section we describe each component of \methodName{} in detail. 

\subsection{Continuous Sign Language Production} \label{sec:continuous_SLP}

To produce continuous sign language sequences from spoken language sentences, we expand the \textit{Progressive Transformer} model proposed by Saunders \etal \cite{saunders2020progressive}. To overcome the issues of under-articulation seen in previous works \cite{saunders2020adversarial,saunders2020progressive}, we use a \acf{mdn} \cite{bishop1994mixture} to model the variation found in sign language, as seen on the left of Figure \ref{fig:model_overview}. Multiple distributions are used to parameterise the entire prediction subspace, with each mixture component modelling a separate valid movement into the future. Formally, given a source sequence, $x_{1:U}$, we can model the conditional probability of producing a sign pose, $y_{t}$, as:
\begin{equation} \label{eq:MDN_conditional_probability}
    p(y_{t}|x_{1:U}) = \sum_{i=1}^{M} \alpha_{i}(x_{1:U}) \phi_{i}(y_{t}|x_{1:U}) 
\end{equation}
where $M$ is the number of mixture components in the \ac{mdn}. $\alpha_{i}(x_{1:U})$ is the mixture weight of the $i^{th}$ distribution, regarded as a prior probability of the sign pose being generated from this mixture component. $\phi_{i}(y_{t}|x_{1:U}) $ is the conditional density of the sign pose for the $i^{th}$ mixture, which can be expressed as a Gaussian distribution:
\begin{equation} \label{eq:MDN_gaussian_dist}
    \phi_{i}(y_{t}|x_{1:U}) = \frac{1} { \sigma_{i}(x_{1:U}) \sqrt{2\pi} } 
\; exp^{ \frac{\left \| y_{t} - \mu_{i}(x_{1:U})  \right \| ^2}{2 \sigma_{i}(x_{1:U})^2} }
\end{equation}
where $\mu_{i}(x_{1:U})$ and $\sigma_{i}(x_{1:U})$ denote the mean and variance of the $i^{th}$ distribution. During training, we minimise the negative log likelihood of the ground truth data coming from our predicted mixture distribution. At inference, the full sign pose sequence, $Y=y_{1:T}$, can be sampled from the mixture distribution in an auto-regressive manner. 

To condition our signer generation model, we convert the sign pose sequence into a heat-map representation using a pose-conditioning layer. Specifically, each limb is plotted on a separate feature channel, where a limb is defined as the connecting line between two corresponding joints. This results in a sequence of pose-conditioned features that each represents a video frame.

\subsection{Photo-Realistic Sign Language Generation}

To generate a photo-realistic sign language video, $Z$, conditioned on the produced sign language pose sequence, $Y$, we propose a method for video-to-video signer generation. Taking inspiration from \cite{chan2019everybody}, in the conditional \ac{gan} setup, a generator network, $G$, competes in a min-max game against a multi-scale discriminator, $D = (D_{1}, D_{2}, D_{3})$. The goal of $G$ is to synthesise images of similar quality to ground-truth images, in order to fool $D$. Conversely, the aim of $D$ is to discern the ``fake'' images from the ``real'' images. For our purposes, $G$ synthesises images of a signer given a human pose, $y_{t}$, and a style image, $I^{S}$, as shown on the right of Figure \ref{fig:model_overview}.

Following \cite{isola2017image}, we introduce skip connections to the architecture of $G$ in a \textit{U-Net} structure \cite{ronneberger2015u}. Skip connections propagate pose information across the networks, enabling the generation of fine-grained details. Specifically, we add skip connections between each down-sampling layer $i$ and up-sampling layer $n-i$, where $n$ is the total number of up-sampling layers. 

\subsubsection{Controllable Video Generation}

To enable training on diverse sign language datasets, we propose a style-controllable video generation approach. A style image, $I^{S}$, is provided to condition synthesis alongside the pose sequence, ${(y_{0},...,y_{t})}$, as seen in Figure \ref{fig:model_overview}. During training, the model learns to associate the given style, $S$, with the person-specific aspects of the corresponding target image, $z^{S}_{t}$, such as the clothing or face. The hand pose information is signer invariant and is thus learnt independently from the style, enhancing the quality of the hand synthesis. In Section \ref{sec:quant_experiments} we show the effect of training on a larger, multi-signer dataset compared to a smaller single signer corpus.

Controllable generation enables \methodName{} to make use of the variability in signer appearance in the data. A multi-modal distribution of sign language videos in different styles, $Z^{S}$, can be produced, where \hbox{$S \in \{0,N_{S}\}$} represents the styles seen during training. Furthermore, given a few examples of an unseen signer appearance, our model can be fine-tuned to generate a new style not seen at training, with a consistent synthesis of appearance-invariant aspects such as hands. The proposed controllable generation enables diversity in signer generation, which has been highlighted as important by Deaf focus groups \cite{kipp2011assessing}. Figure \ref{fig:qual_examples} provides examples of varying signer appearance generation.

\subsubsection{Hand Keypoint Loss}

\begin{figure}[t!]
    \centering
    \includegraphics[width=1.00 \linewidth]{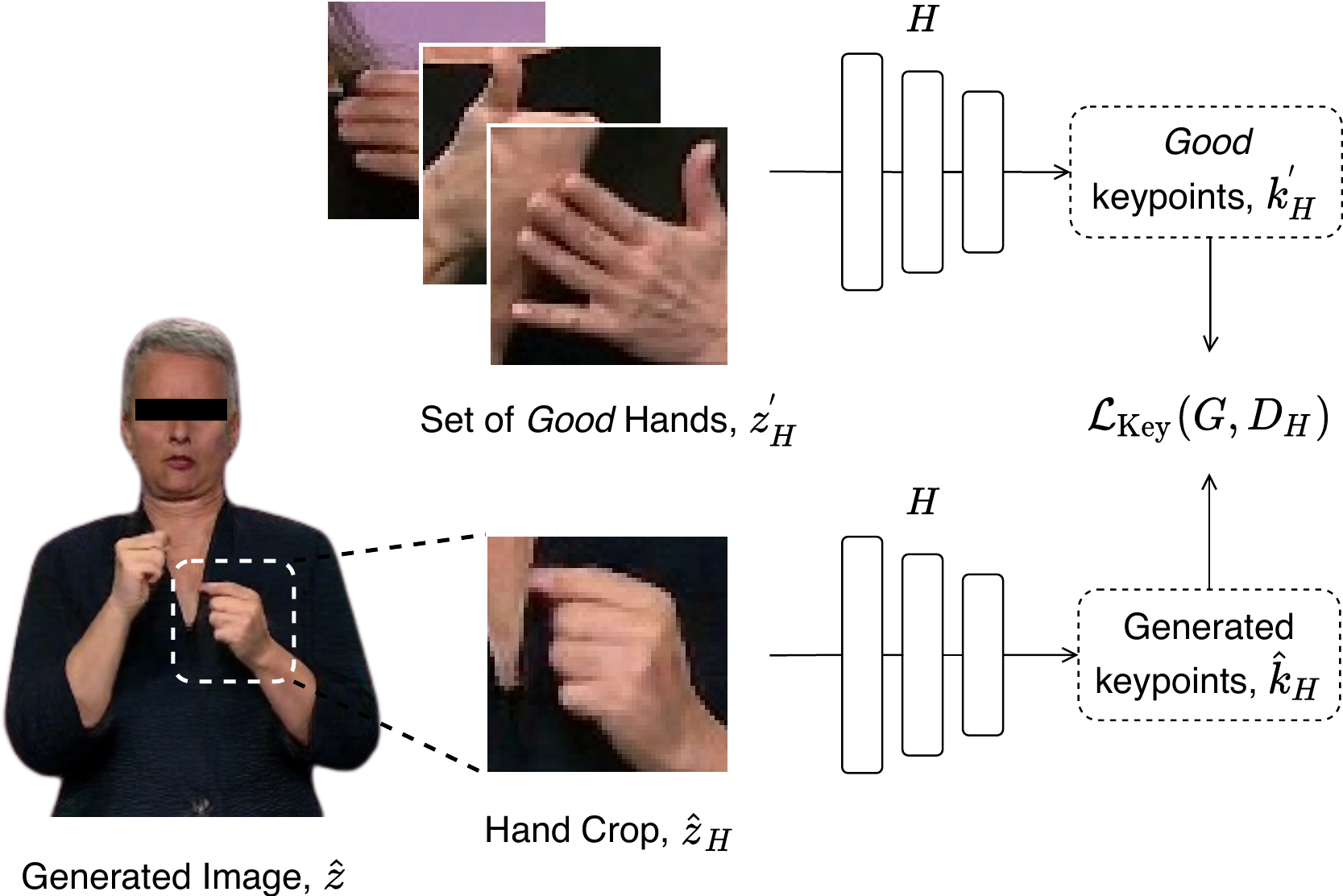}
    \caption{Hand keypoint loss overview. A keypoint discriminator, $D_{H}$, compares keypoints extracted from generated hands, $\hat{k}_{H}$ and sampled \textit{good} hands, ${k}^{'}_{H}$.}
    \label{fig:keypoint_loss}
\end{figure}%

\begin{figure*}[t!]
    \centering
    \includegraphics[width=0.99 \linewidth]{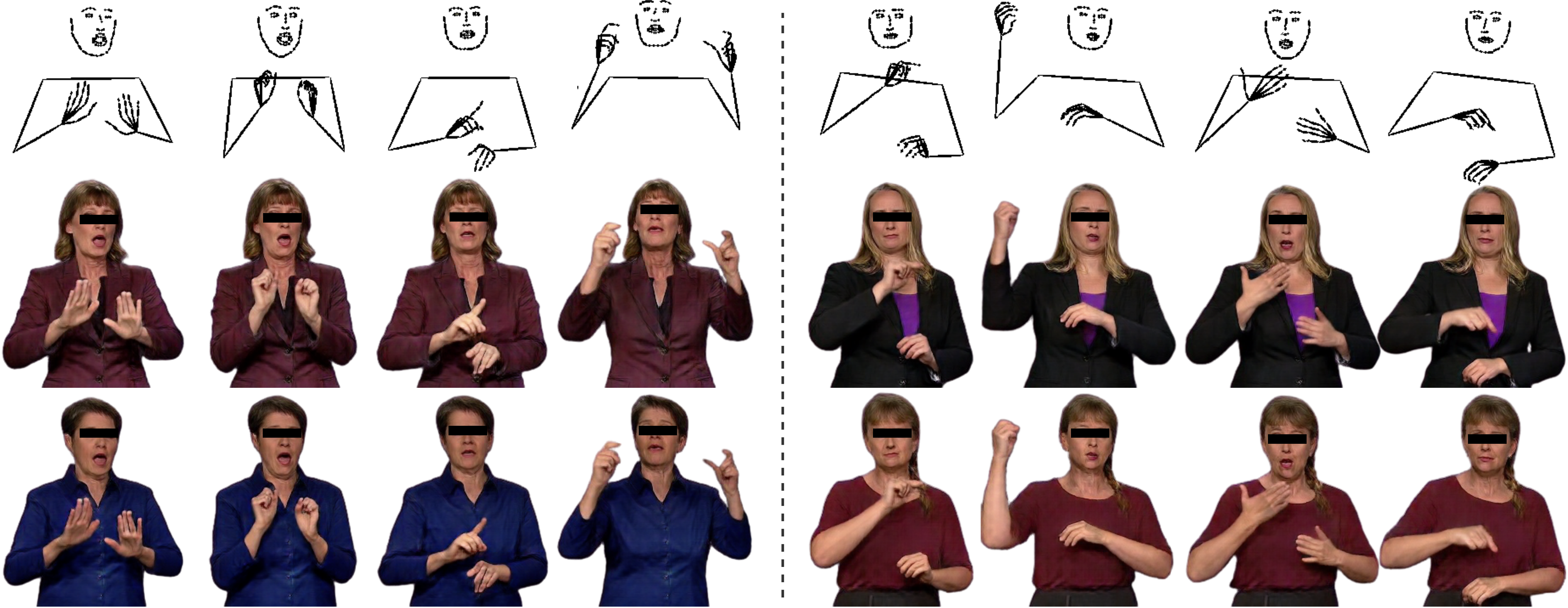}
    \caption{Photo-realistic sign language video generation examples, highlighting the variability in signer appearance.}
    \label{fig:qual_examples}
\end{figure*}%

Previous pose-conditioned human synthesis methods have failed to generate realistic and accurate hand images. To enhance the quality of hand synthesis, we introduce a novel loss that operates in the keypoint space, as shown in Figure \ref{fig:keypoint_loss}. A pre-trained 2D hand pose estimator \cite{ge20193d}, $H$, is used to extract hand keypoints, $k_{H}$, from cropped hand regions (\textit{i.e.} a 60x60 patch centered around the middle knuckle), $z_{H}$, as \hbox{$k_{H} = H(z_{H})$}. We avoid operating in the image space due to the existence of blurry hand images in the dataset, whereas the extracted keypoints are invariant to motion-blur. A hand keypoint discriminator, $D_{H}$, then attempts to discern between the ``real'' keypoints, \hbox{$k^{\star}_{H} = H(z_{H})$}, and the ``fake'' keypoints, \hbox{$\hat{k}_{H} = H(G(y_{H}))$}, leading to the objective:
\begin{align}
\label{eq:loss_HandKey}
    \mathcal{L}_{\mathrm{\textsc{Key}}}(G,D_{H}) & =  \mathbb{E}_{y_{H},z_{H}} [\log D_{H}(k^{\star}_{H})] \nonumber \\
    + \mathbb{E}_{y_{H}} & [\log (1-D_{H}(\hat{k}_{H}))]
\end{align}

Furthermore, we collect a set of \textit{good} hands, ${z}^{'}_{H}$, where the extracted keypoints, \hbox{${k}^{'}_{H} = H({z}^{'}_{H})$}, are used to train the hand keypoint discriminator described above. This amends the objective in Equation \ref{eq:loss_HandKey} by replacing $\hat{k}_{H}$ with ${k}^{'}_{H}$, further enhancing the synthesis of hand images. The proposed hand keypoint loss allows the use of training samples from multiple datasets, as it is appearance invariant and removes background artifacts. Figure \ref{fig:qual_examples} showcases qualitative examples of our high-quality hand synthesis.

\begin{table*}[t!]
\centering
\resizebox{0.83\linewidth}{!}{%
\begin{tabular}{@{}p{2.8cm}ccccc|ccccc@{}}
\toprule
     & \multicolumn{5}{c}{DEV SET}  & \multicolumn{5}{c}{TEST SET} \\ 
\multicolumn{1}{c|}{Approach:}  & BLEU-4         & BLEU-3         & BLEU-2         & BLEU-1         & ROUGE          & BLEU-4         & BLEU-3         & BLEU-2         & BLEU-1         & ROUGE          \\ \midrule
\multicolumn{1}{r|}{Progressive Transformers \cite{saunders2020progressive}}    & 11.82 & 14.80 & 19.97 & 31.41 & 33.18 & 10.51 & 13.54 & 19.04 & 31.36 & 32.46 \\ 
\multicolumn{1}{r|}{Adversarial Training \cite{saunders2020adversarial}} & \textbf{12.65} & \textbf{15.61} & \textbf{20.58} & \textbf{31.84} & \textbf{33.68} & 10.81 & 13.72 & 18.99 & 30.93 & 32.74 \\

\multicolumn{1}{r|}{\textbf{Mixture Density Networks (Ours)}} & 11.54 & 14.48 & 19.63 & 30.94 & 33.40 & 11.68 & 14.55 & \textbf{19.70} & \textbf{31.56} & \textbf{33.19} \\ 
\multicolumn{1}{r|}{\textbf{\methodName{} (Ours)}} & 11.74 & 14.38 & 18.81 & 27.07 & 27.83 & \textbf{12.18} & \textbf{14.84} & 19.26 & 27.63 & 29.05 \\
\bottomrule
\end{tabular}%
}
\caption{Back translation results on the \ac{ph14t} dataset for the \textit{Text to Pose} task.}
\label{tab:back_translation}
\end{table*}

\subsubsection{Full Objective}

In standard image-to-image translation frameworks \cite{isola2017image,wang2018high}, $G$ is trained using a combination of adversarial and perceptual losses. We update the multi-scale adversarial loss, $\mathcal{L}_{GAN}(G,D)$, to reflect our controllable generation with a joint conditioning on sign pose, $y_{t}$, and style image, $I^{S}$:
\begin{align}
\label{eq:loss_gan}
    \mathcal{L}_{GAN}(G,D) & = \sum^{k}_{i=1} \;  \mathbb{E}_{y_{t},z_{t}} [\log D_{i}(z_{t} \mid y_{t},I^{S})] \nonumber \\
    + \mathbb{E}_{y_{t}} [\log & (1-D_{i}(G(y_{t},I^{S}) \mid y_{t},I^{S}))]  
\end{align}
\noindent where $k = 3$ reflects the multi-scale discriminator. The adversarial loss is supplemented with two feature-matching losses; $\mathcal{L}_{FM}(G,D)$, the discriminator feature-matching loss presented in pix2pixHD \cite{wang2018high}, and $\mathcal{L}_{VGG}(G,D)$, the perceptual reconstruction loss \cite{johnson2016perceptual} which compares pretrained VGGNet \cite{simonyan2014very} features at multiple layers of the network. We adapt our model to the video domain with the inclusion of a temporal consistency loss, $\mathcal{L}_{\mathrm{T}} (G) = (\hat{\delta} - \delta^{\star})^2$, where $\hat{\delta}$ and $\delta^{\star}$ are the pixel-wise frame differences for produced and ground truth data respectively. Our full objective, $\mathcal{L}_{Total}$, is a weighted sum of these, alongside our proposed hand keypoint loss (Eq. \ref{eq:loss_HandKey}), as:
\begin{align}
\label{eq:loss_FM}
    \mathcal{L}_{Total} & = 
    \min_{G} (( \max_{D_{i}}  \sum^{k}_{i=1} \; \mathcal{L}_{GAN}(G,D_{i}))  \nonumber  \\
     + \lambda_{FM} & \sum^{k}_{i=1} \; \mathcal{L}_{FM}(G,D_{i}) 
     + \lambda_{VGG} \; \mathcal{L}_{VGG} (G(y_{t},I^{S}),z_{t}) \nonumber  \\
     + \lambda_{\textsc{Key}} & \; \mathcal{L}_{\mathrm{\textsc{Key}}}(G,D_{H}) 
     + \lambda_{T} \; \mathcal{L}_{T} (G) )
\end{align}
\noindent where $k = 3$ and $\lambda_{FM},\lambda_{VGG},\lambda_{\textsc{Key}},\lambda_{T}$ weight the contributions of each loss.

\section{Experiments} \label{sec:experiments}

In this section, we evaluate the performance of \methodName{}. We first outline our experimental setup then perform quantitative, qualitative and user evaluation. 

\subsection{Experimental Setup}

\paragraph{Datasets}

We use the challenging \ac{ph14t} dataset released by Camgoz \etal \cite{camgoz2018neural} to train our continuous \ac{slp} model, with setup as proposed in Saunders \etal \cite{saunders2020progressive}. We train our human synthesis model with a collected corpus of high-quality sign language interpreter broadcast data. We use separate datasets for the two network components to overcome the limitations of each dataset. The video quality of \ac{ph14t} is very low, whereas our collected interpreter data does not have aligned spoken language translations available. We apply the pose normalisation techniques of \cite{chan2019everybody} to transfer between the relevant datasets.

2D upper body joints and facial landmarks are extracted using OpenPose \cite{cao2018openpose}. We use a heat-map representation as pose condition, as described in Section \ref{sec:continuous_SLP}. For the target image, we segment the sign interpreter and replace the background with a consistent colour. We evaluate using an unseen sequence of a signer appearance seen during testing.

\paragraph{Baseline Methods}

We compare the performance of \methodName{} against state-of-the-art image-to-image and video-to-video translation methods, conditioned on skeletal pose images. 1) \textbf{Everybody Dance Now (EDN)} \cite{chan2019everybody} presents a method for ``do as I do'' motion transfer using pose as an intermediate representation. 2) \textbf{Video-to-Video Synthesis (vid2vid)} \cite{wang2018video} uses a spatio-temporal adversarial objective to produce temporally coherent videos. 3) \textbf{Pix2PixHD} \cite{wang2018high} is a high-definition image-to-image translation model, used to generate a video in a frame-by-frame manner. 4) \textbf{Stoll \etal} \cite{stoll2020text2sign} apply pix2pixHD without the VGG loss to produce photo-realistic sign language videos.

\begin{figure*}[t!]
    \centering
    \includegraphics[width=0.95 \linewidth]{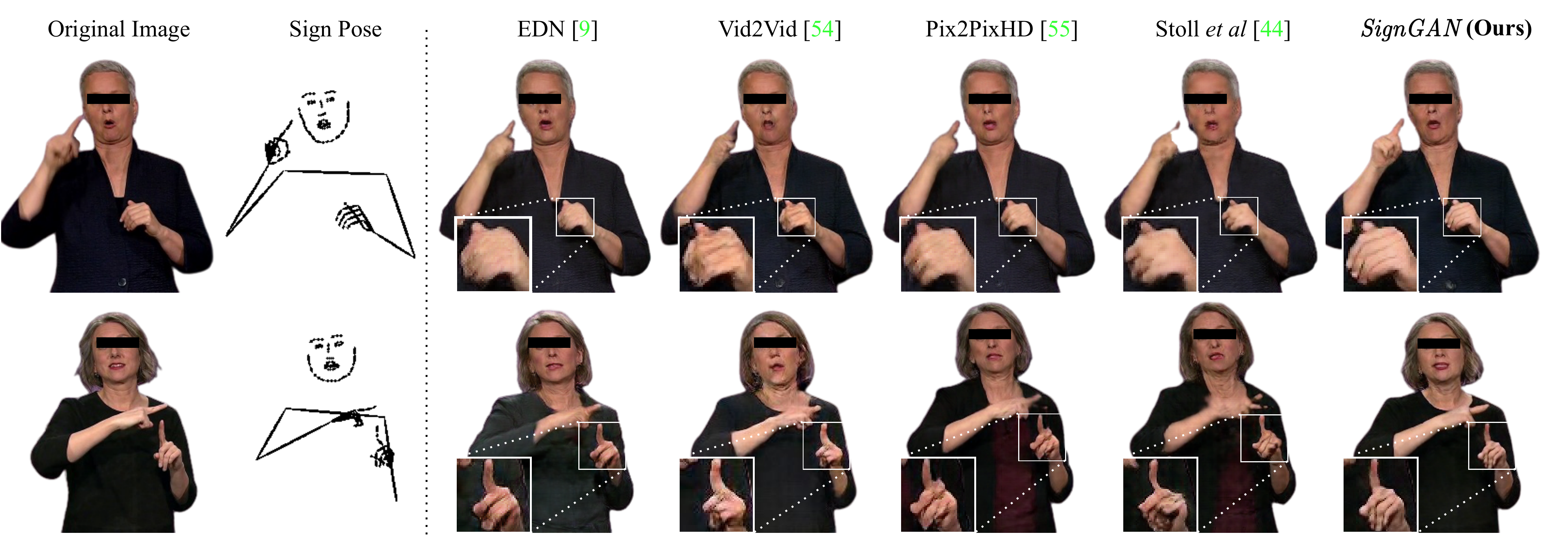}
    \caption{Baseline comparisons trained on a single signer (top) or multiple signer (bottom) dataset. Best viewed in colour.}
    \label{fig:qual_baseline}
\end{figure*}%

\paragraph{Ablation Conditions}

We ablate our proposed model architecture with the following conditions. 1) \textbf{Skip Connections}: We start with the base condition of skip connections added to the \methodName{} architecture. 2) \textbf{Controllable Generation}: In this condition, we include our proposed controllable video generation module. 3) \textbf{Hand Discriminator}: Here we apply a naive discriminator over the generated hand patch. 4) \textbf{Hand Keypoint Loss}: We replace the hand discriminator with our proposed hand keypoint loss. 5) \textbf{\methodName{} (Full)}: Our final condition is the full \methodName{} model with the addition of a hand keypoint loss compared to a set of \textit{good} hands. Each condition is inclusive of all previous additions excluding the hand discriminator.

\paragraph{Evaluation Metrics}

We measure the quality of synthesized images using the following metrics. 1) \textbf{SSIM}: Structural Similarity \cite{wang2004image} over the full image. 2) \textbf{Hand SSIM}: SSIM metric over a 60x60 crop of each hand. 3) \textbf{Hand Pose}: Absolute distance between 2D hand keypoints of the produced and ground truth hand images, using a pre-trained hand pose estimation model \cite{ge20193d}. We note similar metrics were used in prior works \cite{isola2017image,ventura2020can,wang2019few,wang2018video}. 4) \textbf{FID}: Fr\'echet Inception Distance \cite{heusel2017gans} over the full image.

\subsection{Quantitative Evaluation} \label{sec:quant_experiments}

\begin{table}[b!]
\centering
\resizebox{0.99\linewidth}{!}{%
\begin{tabular}{@{}p{0.0cm}cccc@{}}
\toprule
  & \multicolumn{1}{c}{SSIM $\uparrow$} & \multicolumn{1}{c}{Hand SSIM $\uparrow$} & \multicolumn{1}{c}{Hand Pose $\downarrow$} & \multicolumn{1}{c}{FID $\downarrow$} \\ \midrule
\multicolumn{1}{r|}{EDN \cite{chan2019everybody}} & 0.726 & 0.545 & 22.88 & 29.62 \\
\multicolumn{1}{r|}{vid2vid \cite{wang2018video}} & 0.726 & 0.551 & 22.76 & 27.51 \\
\multicolumn{1}{r|}{Pix2PixHD \cite{wang2018high}} & 0.726 & 0.541 & 22.95 & 32.88 \\ 
\multicolumn{1}{r|}{Stoll \etal \cite{stoll2020text2sign}} & 0.733 & 0.547 & 22.81 & 73.38 \\ 
\multicolumn{1}{r|}{\methodName{} (Ours)} & \textbf{0.742} & \textbf{0.588} & \textbf{22.68} & \textbf{24.10} \\
\bottomrule
\end{tabular}%
}
\caption{Baseline model comparison results, on a single signer subset of the data.}
\label{tab:baselines_consistent}
\end{table}

\paragraph{Back Translation}

Our first evaluation is of our continuous \ac{slp} model. The benchmark \ac{slp} evaluation metric is back translation, which uses a pre-trained \ac{slt} model to translate back to spoken language \cite{saunders2020progressive}. We use the state-of-the-art \ac{slt} \cite{camgoz2020sign} as our back translation model, utilizing EfficientNet-B7 \cite{tan2019efficientnet} features as frame representations. Our models are evaluated on the \textit{Text to Pose} task, translating directly from spoken language to sign language video.

We first evaluate our \ac{mdn} formulation for the production of skeleton pose sequences, as an intermediary before photo-realistic video generation. Table \ref{tab:back_translation} shows a performance increase on the test set compared to baseline methods \cite{saunders2020adversarial,saunders2020progressive}. The multimodal modelling of the \ac{mdn} reduces the regression to the mean found in previous deterministic models, leading to a more expressive production.

We next evaluate our full \methodName{} model for photo-realistic sign language video generation. \methodName{} achieves state-of-the-art back translation BLEU-4 performance on the test set, of 12.18. This highlights the increased information content available in photo-realistic videos, \textit{i.e.} a learnt human appearance prior. In addition, this shows the importance of photo-realistic \ac{slp} for sign comprehension.

\paragraph{Baseline Comparisons}

We next evaluate our pose-conditioned signer generation by comparing performance to baselines on the same task, given a sequence of poses as input. For a fair evaluation, we first evaluate using a subset of data containing a single signer appearance. As seen in Table \ref{tab:baselines_consistent}, \methodName{} outperforms all baseline methods for all metrics, particularly for the Hand SSIM score. We believe this is due to the improved quality of synthesized hand images by using the proposed hand keypoint loss.

We next evaluate using a larger, multiple signer dataset in Table \ref{tab:baselines_diverse}, to show the effect of our controllable generation module. \methodName{} again outperforms all baseline methods and achieves a significant performance increase compared to training on a single signer corpus. We believe this is due to the larger variety of hand and body pose present in the diverse dataset, as well as the presence of multiple signers acting as a regularizer. Conversely, baseline methods perform significantly worse for FID scores, due to the lack of ability to control the signer appearance.  

\begin{table}[t!]
\centering
\resizebox{0.99\linewidth}{!}{%
\begin{tabular}{@{}p{0.0cm}cccc@{}}
\toprule
  & \multicolumn{1}{c}{SSIM $\uparrow$} & \multicolumn{1}{c}{Hand SSIM $\uparrow$} & \multicolumn{1}{c}{Hand Pose $\downarrow$} & \multicolumn{1}{c}{FID $\downarrow$} \\ \midrule
\multicolumn{1}{r|}{EDN \cite{chan2019everybody}} & 0.737 & 0.553 & 23.09 & 41.54 \\
\multicolumn{1}{r|}{vid2vid \cite{wang2018video}} & 0.750  & 0.570 & 22.51 & 56.17 \\
\multicolumn{1}{r|}{Pix2PixHD \cite{wang2018high}} & 0.737 & 0.553 & 23.06 & 42.57 \\ 
\multicolumn{1}{r|}{Stoll \etal \cite{stoll2020text2sign}} & 0.727 & 0.533 & 23.17 & 64.01 \\ 
\multicolumn{1}{r|}{\methodName{} (Ours)} & \textbf{0.759} & \textbf{0.605} & \textbf{22.05} & \textbf{27.75}  \\
\bottomrule
\end{tabular}%
}
\caption{Baseline model comparison results, on the full dataset containing multiple signers.}
\label{tab:baselines_diverse}
\end{table}

\paragraph{Ablation Study}

We perform an ablation study of \methodName{} using the multiple signer dataset, with results in Table \ref{tab:ablation}. Our skip connection architecture achieves a strong performance, with SSIM and hand SSIM higher than all baseline models bar vid2vid. The importance of our controllable generation is highlighted by an improvement in SSIM and FID scores when applied. Without this, a generated video will be temporally inconsistent with a blurred appearance.

The hand discriminator performs poorly for both SSIM and hand SSIM, due to the generation of blurred hands. However, our proposed hand keypoint loss improves performance, particularly for hand SSIM, emphasizing the importance of an adversarial loss invariant to blurring. Finally, the full \methodName{} model performs best, particularly for the hand SSIM score. We believe this is due to the increased quality of images used to train our hand keypoint discriminator, prompting an enhanced synthesis.

\begin{figure*}[t!]
    \centering
    \includegraphics[width=0.97 \linewidth]{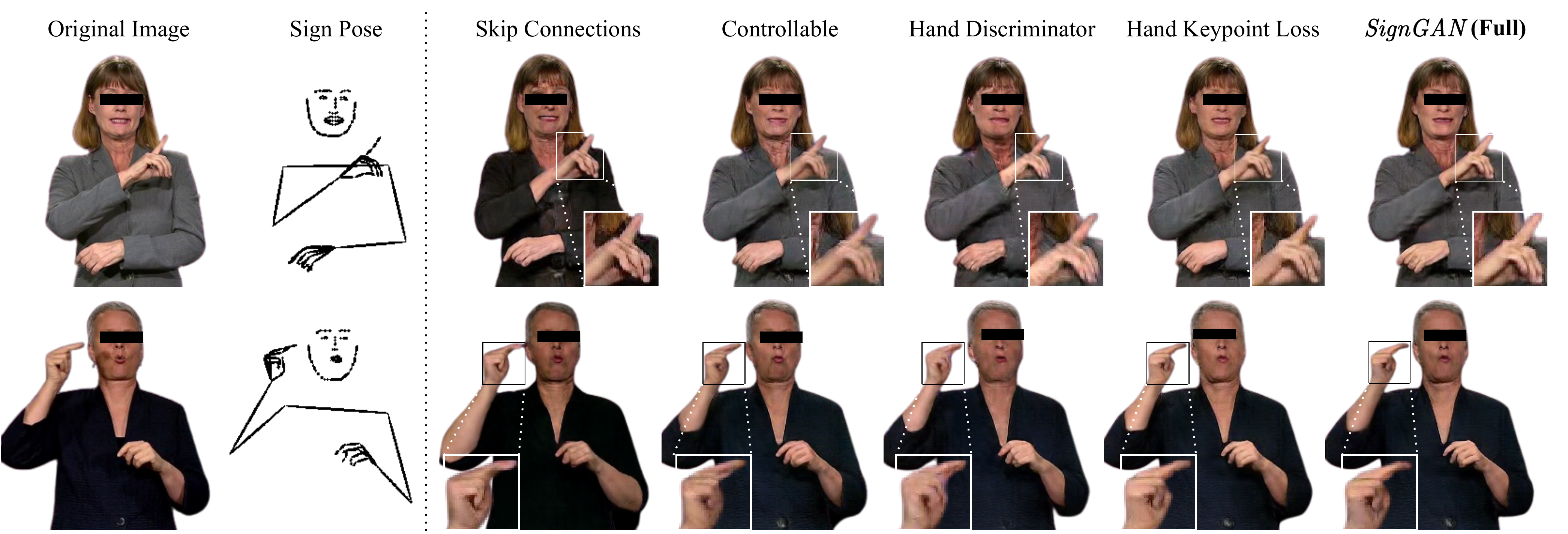}
    \caption{Ablation Study of \methodName{} trained on the multiple signer dataset. Best viewed in colour.}
    \label{fig:qual_ablation}
\end{figure*}%

\begin{table}[t!]
\centering
\resizebox{0.99\linewidth}{!}{%
\begin{tabular}{@{}p{0.0cm}ccccc@{}}
\toprule
  & \multicolumn{1}{c}{SSIM $\uparrow$} & \multicolumn{1}{c}{Hand SSIM $\uparrow$} & \multicolumn{1}{c}{Hand Pose $\downarrow$} & \multicolumn{1}{c}{FID $\downarrow$} \\ \midrule
\multicolumn{1}{r|}{Skip Connections} & 0.743 & 0.582 & 22.87 & 39.33 \\
\multicolumn{1}{r|}{Controllable Gen.} & 0.752 & 0.587 & 22.09 & 32.22 \\
\multicolumn{1}{r|}{Hand Discriminator} & 0.738 & 0.565 & 22.81 & 39.22 \\
\multicolumn{1}{r|}{Hand Keypoint Loss} & 0.758 & 0.598 & \textbf{21.89} & 29.02 \\
\multicolumn{1}{r|}{\methodName{} (Full)} & \textbf{0.759} & \textbf{0.605} & 22.05 & \textbf{27.75} \\
\bottomrule
\end{tabular}%
}
\caption{Ablation study results, on a multiple signer dataset.}
\label{tab:ablation}
\end{table}

\begin{table}[b!]
\centering
\resizebox{0.90\linewidth}{!}{%
\begin{tabular}{@{}p{0.0cm}cccc@{}}
\toprule
& \multicolumn{2}{c|}{\;\; Single Signer \;\;} &  \multicolumn{2}{c}{\;\; Multiple Signer \;\;}  \\ 
 & \multicolumn{1}{c}{Body} & \multicolumn{1}{c|}{Hand} & \multicolumn{1}{c}{Body} & \multicolumn{1}{c}{Hand}  \\ \midrule
\multicolumn{1}{r|}{EDN \cite{chan2019everybody}} & 97.8\% & 97.8\% & 100\% & 97.8\% \\
\multicolumn{1}{r|}{vid2vid \cite{wang2018video}} & 97.8\% & 95.6\% & 85.9\% & 84.8\% \\
\multicolumn{1}{r|}{Pix2PixHD \cite{wang2018high}} & 96.7\% & 96.7\% & 98.9\% & 100\% \\
\multicolumn{1}{r|}{Stoll \etal \cite{stoll2020text2sign}} & 98.9\% & 100\% & 100\% & 100\% \\
\bottomrule
\end{tabular}%
}
\caption{Perceptual study results, showing the percentage of participants who preferred \textbf{our} method to the baseline trained on a single or multiple signer dataset.}
\label{tab:perceptual}
\end{table}

\paragraph{Perceptual Study}

We perform a perceptual study of our video generation, showing participants pairs of 10 second videos generated by \methodName{} and a corresponding baseline method. We generated videos using either a single or multiple signer dataset. Participants were asked to select which video was more visually realistic, with a separate focus on the body and hands. In total, 46 participant completed the study, of which 28\% were signers, each viewing 2 randomly selected videos from each of the baselines. Table \ref{tab:perceptual} shows the percentage of participants who preferred the outputs of \methodName{} to the baseline method.

It can be clearly seen that our outputs were preferred by participants compared to all baseline models for both body (97.3\% average) and hand (96.5\% average) synthesis. Vid2vid was the strongest contender, with our productions preferred only 91\% of the time, compared to 99.7\% for Stoll \etal. In addition, \methodName{} was preferred when the models were trained on the multiple signer dataset, highlighting the effect of our controllable video generation module.

\subsection{Qualitative Evaluation} \label{sec:qual_experiments}

\paragraph{Baseline Comparison}

As seen in the baseline comparisons of Figure \ref{fig:qual_baseline}, \methodName{} clearly generates the most natural-looking hands. Other approaches generate hands that are blurred, due to the existence of motion blur in the training data and the lack of a hand-specific loss. When trained on the larger, multiple signer dataset (bottom), baselines struggle to generate a consistent signer appearance. In contrast, \methodName{} generates an identical style to the original image, due to the proposed controllable video generation. Although superior results can be seen in image synthesis, we believe more significant differences can be seen in the video outputs. We recommend the reader to view the comparison videos provided in supplementary material. 

\paragraph{Ablation Study}

\begin{figure}[b!]
    \centering
    \includegraphics[width=0.99 \linewidth]{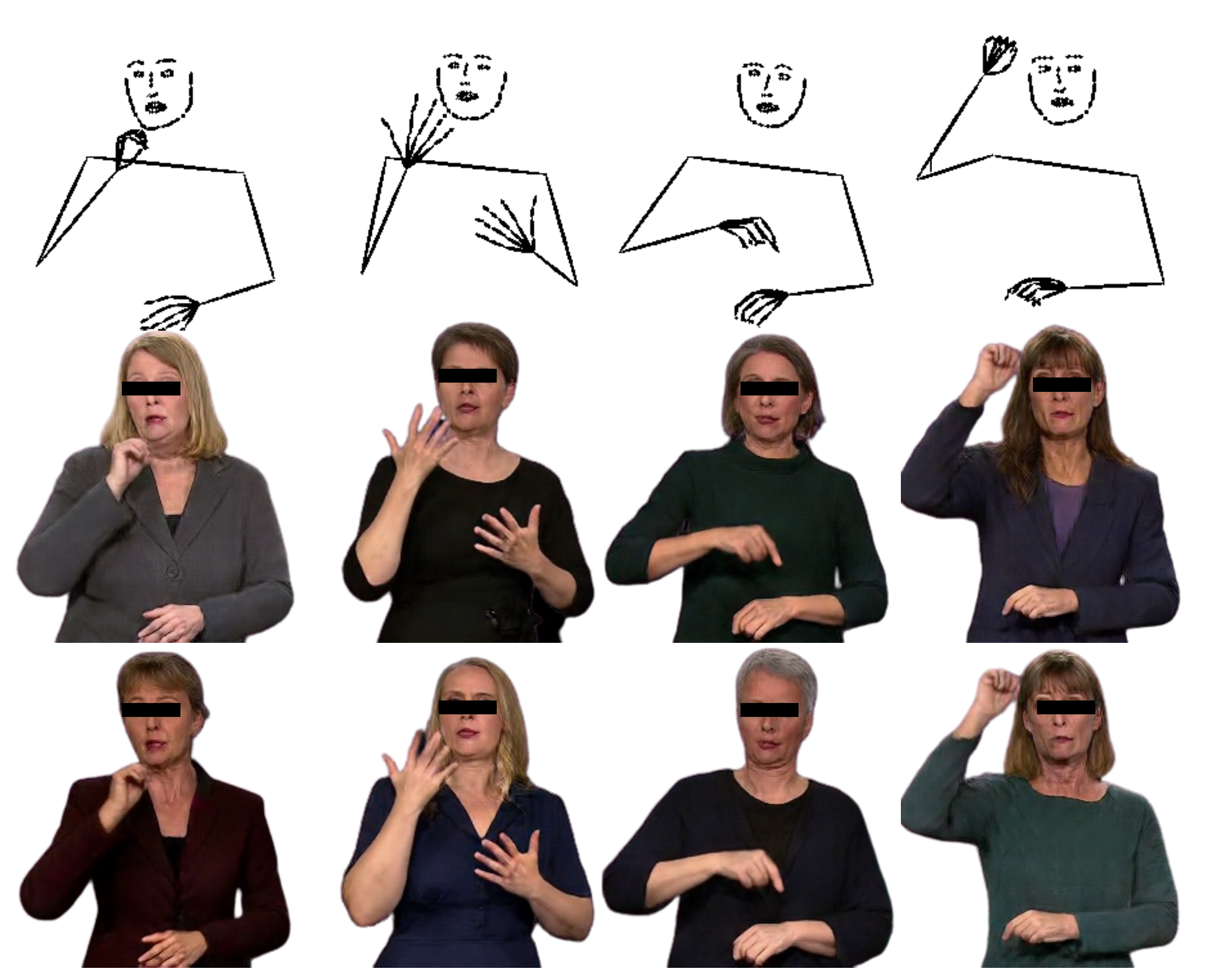}
    \caption{Variable generation of 8 signer appearances.}
    \label{fig:controllable_examples}
\end{figure}

Figure \ref{fig:qual_ablation} shows qualitative results when ablating our model. The effect of the proposed controllable video generation is highlighted, as without it the signer appearance is inconsistent. The classic hand discriminator can be seen to produce blurry hands due to the existence of motion blur in the training dataset. Our proposed hand keypoint loss generates higher quality hands with considerable detail and sharper synthesis, particularly when trained using a set of \textit{good} hands in the \methodName{} (full) setup. Although the effect is subtle, this is an important addition for the understanding of sign language videos, where the focus is often on the manual features. Figure \ref{fig:controllable_examples} showcases the full repertoire of interpreters generated from a single model, showing the control available to the viewer.

\section{Conclusion} \label{sec:conc}

\acf{slp} requires the generation of photo-realistic sign language videos to be understandable by the Deaf \cite{ventura2020can}, whereas prior work has produced skeleton pose sequences. In this paper, we proposed \methodName{}, the first \ac{slp} model to produce photo-realistic continuous sign language videos directly from spoken language. We employed a transformer architecture with a \acf{mdn} formulation to translate from spoken language to skeletal pose. Pose sequences are subsequently used to condition the generation of a photo-realistic sign language video using a human synthesis module.

We proposed a novel keypoint-based loss function to significantly improve the quality of hand synthesis, operating in the keypoint space to avoid issues caused by motion blur. Additionally, we proposed a method for controllable video generation, enabling training on large, diverse sign language datasets and providing the ability to control the signer appearance at inference. Finally, we collected a varied dataset of sign language interpreters and showed that \methodName{} outperforms all baseline methods for quantitative evaluation and human perceptual studies.

\section{Acknowledgements}

This work received funding from the SNSF Sinergia project 'SMILE' (CRSII2 160811), the European Union's Horizon2020 research and innovation programme under grant agreement no. 762021 'Content4All' and the EPSRC project 'ExTOL' (EP/R03298X/1). This work reflects only the authors view and the Commission is not responsible for any use that may be made of the information it contains. We would also like to thank NVIDIA Corporation for their GPU grant. We would like to thank SWISSTXT for access to the broadcast footage.

{\small
\bibliographystyle{ieee_fullname}
\bibliography{bibliography}

\begin{thebibliography}{10}\itemsep=-1pt

\bibitem{antonakos2012unsupervised}
Epameinondas Antonakos, Vassilis Pitsikalis, Isidoros Rodomagoulakis, and
  Petros Maragos.
\newblock {Unsupervised Classification of Extreme Facial Events using Active
  Appearance Models Tracking for Sign Language Videos}.
\newblock In {\em 19th IEEE International Conference on Image Processing
  (ICIP)}, 2012.

\bibitem{balakrishnan2018synthesizing}
Guha Balakrishnan, Amy Zhao, Adrian~V Dalca, Fredo Durand, and John Guttag.
\newblock {Synthesizing Images of Humans in Unseen Poses}.
\newblock In {\em Proceedings of the IEEE Conference on Computer Vision and
  Pattern Recognition (CVPR)}, 2018.

\bibitem{bauer2000video}
Britta Bauer, Hermann Hienz, and K-F Kraiss.
\newblock {Video-Based Continuous Sign Language Recognition using Statistical
  Methods}.
\newblock In {\em Proceedings of International Conference on Pattern
  Recognition (ICPR)}, 2000.

\bibitem{bishop1994mixture}
Christopher~M Bishop.
\newblock {Mixture Density Networks}.
\newblock {\em Technical Report, Citeseer}, 1994.

\bibitem{camgoz2017subunets}
Necati~Cihan Camgoz, Simon Hadfield, Oscar Koller, and Richard Bowden.
\newblock {SubUNets: End-to-end Hand Shape and Continuous Sign Language
  Recognition}.
\newblock In {\em Proceedings of the IEEE International Conference on Computer
  Vision (ICCV)}, 2017.

\bibitem{camgoz2018neural}
Necati~Cihan Camgoz, Simon Hadfield, Oscar Koller, Hermann Ney, and Richard
  Bowden.
\newblock {Neural Sign Language Translation}.
\newblock In {\em Proceedings of the IEEE Conference on Computer Vision and
  Pattern Recognition (CVPR)}, 2018.

\bibitem{camgoz2020multi}
Necati~Cihan Camgoz, Oscar Koller, Simon Hadfield, and Richard Bowden.
\newblock {Multi-channel Transformers for Multi-articulatory Sign Language
  Translation}.
\newblock In {\em Assistive Computer Vision and Robotics Workshop (ACVR)},
  2020.

\bibitem{cao2018openpose}
Zhe Cao, Gines Hidalgo, Tomas Simon, Shih-En Wei, and Yaser Sheikh.
\newblock {OpenPose: Realtime Multi-Person 2D Pose Estimation using Part
  Affinity Fields}.
\newblock In {\em Proceedings of the IEEE Conference on Computer Vision and
  Pattern Recognition (CVPR)}, 2017.

\bibitem{chan2019everybody}
Caroline Chan, Shiry Ginosar, Tinghui Zhou, and Alexei~A Efros.
\newblock {Everybody Dance Now}.
\newblock In {\em Proceedings of the IEEE International Conference on Computer
  Vision (CVPR)}, 2019.

\bibitem{camgoz2020sign}
Necati Cihan~Camgoz, Oscar Koller, Simon Hadfield, and Richard Bowden.
\newblock {Sign Language Transformers: Joint End-to-end Sign Language
  Recognition and Translation}.
\newblock In {\em Proceedings of the IEEE Conference on Computer Vision and
  Pattern Recognition (CVPR)}, 2020.

\bibitem{cox2002tessa}
Stephen Cox, Michael Lincoln, Judy Tryggvason, Melanie Nakisa, Mark Wells,
  Marcus Tutt, and Sanja Abbott.
\newblock {TESSA, a System to Aid Communication with Deaf People}.
\newblock In {\em Proceedings of the ACM International Conference on Assistive
  Technologies}, 2002.

\bibitem{cui2017recurrent}
Runpeng Cui, Hu Liu, and Changshui Zhang.
\newblock {Recurrent Convolutional Neural Networks for Continuous Sign Language
  Recognition by Staged Optimization}.
\newblock In {\em Proceedings of the IEEE Conference on Computer Vision and
  Pattern Recognition (CVPR)}, 2017.

\bibitem{deng2020disentangled}
Yu Deng, Jiaolong Yang, Dong Chen, Fang Wen, and Xin Tong.
\newblock {Disentangled and Controllable Face Image Generation via 3D
  Imitative-Contrastive Learning}.
\newblock In {\em Proceedings of the IEEE Conference on Computer Vision and
  Pattern Recognition (CVPR)}, 2020.

\bibitem{ferreira2020learning}
Jo{\~a}o~P Ferreira, Thiago~M Coutinho, Thiago~L Gomes, Jos{\'e}~F Neto, Rafael
  Azevedo, Renato Martins, and Erickson~R Nascimento.
\newblock {Learning to Dance: A Graph Convolutional Adversarial Network to
  Generate Rrealistic Dance Motions from Audio}.
\newblock {\em Computers \& Graphics}, 2020.

\bibitem{forster2014extensions}
Jens Forster, Christoph Schmidt, Oscar Koller, Martin Bellgardt, and Hermann
  Ney.
\newblock {Extensions of the Sign Language Recognition and Translation Corpus
  RWTH-PHOENIX-Weather}.
\newblock In {\em Proceedings of the International Conference on Language
  Resources and Evaluation (LREC)}, 2014.

\bibitem{ge20193d}
Liuhao Ge, Zhou Ren, Yuncheng Li, Zehao Xue, Yingying Wang, Jianfei Cai, and
  Junsong Yuan.
\newblock {3D Hand Shape and Pose Estimation from a Single RGB Image}.
\newblock In {\em Proceedings of the IEEE Conference on Computer Vision and
  Pattern Recognition (CVPR)}, 2019.

\bibitem{goodfellow2014generative}
Ian Goodfellow, Jean Pouget-Abadie, Mehdi Mirza, Bing Xu, David Warde-Farley,
  Sherjil Ozair, Aaron Courville, and Yoshua Bengio.
\newblock {Generative Adversarial Nets}.
\newblock In {\em Proceedings of the Advances in Neural Information Processing
  Systems (NIPS)}, 2014.

\bibitem{grobel1997isolated}
Kirsti Grobel and Marcell Assan.
\newblock {Isolated Sign Language Recognition using Hidden Markov Models}.
\newblock In {\em IEEE International Conference on Systems, Man, and
  Cybernetics}, 1997.

\bibitem{heusel2017gans}
Martin Heusel, Hubert Ramsauer, Thomas Unterthiner, Bernhard Nessler, and Sepp
  Hochreiter.
\newblock {GANs Trained by a Two Time-Scale Update Rule Converge to a Local
  Nash Equilibrium}.
\newblock In {\em Proceedings of the Advances in Neural Information Processing
  Systems (NIPS)}, 2017.

\bibitem{isola2017image}
Phillip Isola, Jun-Yan Zhu, Tinghui Zhou, and Alexei~A Efros.
\newblock {Image-to-Image Translation with Conditional Adversarial Networks}.
\newblock In {\em Proceedings of the IEEE Conference on Computer Vision and
  Pattern Recognition (CVPR)}, 2017.

\bibitem{johnson2016perceptual}
Justin Johnson, Alexandre Alahi, and Li Fei-Fei.
\newblock {Perceptual Losses for Real-Time Style Transfer and
  Super-Resolution}.
\newblock In {\em Proceedings of the European Conference on Computer Vision
  (ECCV)}, 2016.

\bibitem{karpouzis2007educational}
Kostas Karpouzis, George Caridakis, S-E Fotinea, and Eleni Efthimiou.
\newblock {Educational Resources and Implementation of a Greek Sign Language
  Synthesis Architecture}.
\newblock {\em Computers \& Education (CAEO)}, 2007.

\bibitem{kipp2011assessing}
Michael Kipp, Quan Nguyen, Alexis Heloir, and Silke Matthes.
\newblock {Assessing the Deaf User Perspective on Sign Language Avatars}.
\newblock In {\em The Proceedings of the 13th International ACM SIGACCESS
  Conference on Computers and Accessibility (ASSETS)}, 2011.

\bibitem{ko2019neural}
Sang-Ki Ko, Chang~Jo Kim, Hyedong Jung, and Choongsang Cho.
\newblock {Neural Sign Language Translation based on Human Keypoint
  Estimation}.
\newblock {\em Applied Sciences}, 2019.

\bibitem{koller2019weakly}
Oscar Koller, Necati~Cihan Camgoz, Richard Bowden, and Hermann Ney.
\newblock {Weakly Supervised Learning with Multi-Stream CNN-LSTM-HMMs to
  Discover Sequential Parallelism in Sign Language Videos}.
\newblock {\em {IEEE Transactions on Pattern Analysis and Machine Intelligence
  (TPAMI)}}, 2019.

\bibitem{koller2015continuous}
Oscar Koller, Jens Forster, and Hermann Ney.
\newblock {Continuous Sign Language Recognition: Towards Large Vocabulary
  Statistical Recognition Systems Handling Multiple Signers}.
\newblock {\em Computer Vision and Image Understanding (CVIU)}, 2015.

\bibitem{kowalski2020config}
Marek Kowalski, Stephan~J Garbin, Virginia Estellers, Tadas Baltru{\v{s}}aitis,
  Matthew Johnson, and Jamie Shotton.
\newblock {CONFIG: Controllable Neural Face Image Generation}.
\newblock In {\em Proceedings of the European Conference on Computer Vision
  (ECCV)}, 2020.

\bibitem{liu2019gesture}
Yahui Liu, Marco De~Nadai, Gloria Zen, Nicu Sebe, and Bruno Lepri.
\newblock {Gesture-to-Gesture Translation in the Wild via Category-Independent
  Conditional Maps}.
\newblock In {\em Proceedings of the 27th ACM International Conference on
  Multimedia}, 2019.

\bibitem{ma2017pose}
Liqian Ma, Xu Jia, Qianru Sun, Bernt Schiele, Tinne Tuytelaars, and Luc
  Van~Gool.
\newblock {Pose Guided Person Image Generation}.
\newblock In {\em Advances in Neural Information Processing Systems (NIPS)},
  2017.

\bibitem{mallya2020world}
Arun Mallya, Ting-Chun Wang, Karan Sapra, and Ming-Yu Liu.
\newblock {World-Consistent Video-to-Video Synthesis}.
\newblock In {\em Proceedings of the European Conference on Computer Vision
  (ECCV)}, 2020.

\bibitem{mcdonald2016automated}
John McDonald, Rosalee Wolfe, Jerry Schnepp, Julie Hochgesang, Diana~Gorman
  Jamrozik, Marie Stumbo, Larwan Berke, Melissa Bialek, and Farah Thomas.
\newblock {Automated Technique for Real-Time Production of Lifelike Animations
  of American Sign Language}.
\newblock {\em Universal Access in the Information Society (UAIS)}, 2016.

\bibitem{orbay2020neural}
Alptekin Orbay and Lale Akarun.
\newblock {Neural Sign Language Translation by Learning Tokenization}.
\newblock In {\em IEEE International Conference on Automatic Face and Gesture
  Recognition (FG)}, 2020.

\bibitem{ozdemir2016isolated}
O{\u{g}}ulcan {\"O}zdemir, Necati~Cihan Camg{\"o}z, and Lale Akarun.
\newblock {Isolated Sign Language Recognition using Improved Dense
  Trajectories}.
\newblock In {\em Proceedings of the Signal Processing and Communication
  Application Conference (SIU)}, 2016.

\bibitem{radford2015unsupervised}
Alec Radford, Luke Metz, and Soumith Chintala.
\newblock {Unsupervised Representation Learning with Deep Convolutional
  Generative Adversarial Networks}.
\newblock {\em arXiv preprint arXiv:1511.06434}, 2015.

\bibitem{ren2020self}
Xuanchi Ren, Haoran Li, Zijian Huang, and Qifeng Chen.
\newblock {Self-Supervised Dance Video Synthesis Conditioned on Music}.
\newblock In {\em Proceedings of the 28th ACM International Conference on
  Multimedia}, 2020.

\bibitem{ronneberger2015u}
Olaf Ronneberger, Philipp Fischer, and Thomas Brox.
\newblock {U-net: Convolutional Networks for Biomedical Image Segmentation}.
\newblock In {\em International Conference on Medical Image Computing and
  Computer-Assisted Intervention (MIC-CAI))}, 2015.

\bibitem{saunders2020adversarial}
Ben Saunders, Necati~Cihan Camgoz, and Richard Bowden.
\newblock {Adversarial Training for Multi-Channel Sign Language Production}.
\newblock In {\em Proceedings of the British Machine Vision Conference (BMVC)},
  2020.

\bibitem{saunders2020progressive}
Ben Saunders, Necati~Cihan Camgoz, and Richard Bowden.
\newblock {Progressive Transformers for End-to-End Sign Language Production}.
\newblock In {\em Proceedings of the European Conference on Computer Vision
  (ECCV)}, 2020.

\bibitem{siarohin2018deformable}
Aliaksandr Siarohin, Enver Sangineto, St{\'e}phane Lathuiliere, and Nicu Sebe.
\newblock {Deformable GANs for Pose-Based Human Image Generation}.
\newblock In {\em Proceedings of the IEEE Conference on Computer Vision and
  Pattern Recognition (CVPR)}, 2018.

\bibitem{simonyan2014very}
Karen Simonyan and Andrew Zisserman.
\newblock {Very Deep Convolutional Networks for Large-Scale Image Recognition}.
\newblock {\em arXiv preprint arXiv:1409.1556}, 2014.

\bibitem{starner1997real}
Thad Starner and Alex Pentland.
\newblock {Real-time American Sign Language Recognition from Video using Hidden
  Markov Models}.
\newblock {\em Motion-Based Recognition}, 1997.

\bibitem{stokoe1980sign}
William~C Stokoe.
\newblock {Sign Language Structure}.
\newblock {\em Annual Review of Anthropology}, 1980.

\bibitem{stoll2018sign}
Stephanie Stoll, Necati~Cihan Camgoz, Simon Hadfield, and Richard Bowden.
\newblock {Sign Language Production using Neural Machine Translation and
  Generative Adversarial Networks}.
\newblock In {\em Proceedings of the British Machine Vision Conference (BMVC)},
  2018.

\bibitem{stoll2020text2sign}
Stephanie Stoll, Necati~Cihan Camgoz, Simon Hadfield, and Richard Bowden.
\newblock {Text2Sign: Towards Sign Language Production using Neural Machine
  Translation and Generative Adversarial Networks}.
\newblock {\em International Journal of Computer Vision (IJCV)}, 2020.

\bibitem{tamura1988recognition}
Shinichi Tamura and Shingo Kawasaki.
\newblock {Recognition of Sign Language Motion Images}.
\newblock {\em Pattern Recognition}, 1988.

\bibitem{tan2019efficientnet}
Mingxing Tan and Quoc Le.
\newblock {EfficientNet: Rethinking Model Scaling for Convolutional Neural
  Networks}.
\newblock In {\em International Conference on Machine Learning (ICML)}, 2019.

\bibitem{tang2020xinggan}
Hao Tang, Song Bai, Li Zhang, Philip~HS Torr, and Nicu Sebe.
\newblock {XingGAN for Person Image Generation}.
\newblock In {\em Proceedings of the European Conference on Computer Vision
  (ECCV)}, 2020.

\bibitem{tang2018gesturegan}
Hao Tang, Wei Wang, Dan Xu, Yan Yan, and Nicu Sebe.
\newblock {GestureGAN for Hand Gesture-to-Gesture Translation in the wild}.
\newblock In {\em Proceedings of the 26th ACM International Conference on
  Multimedia}, 2018.

\bibitem{tulyakov2018mocogan}
Sergey Tulyakov, Ming-Yu Liu, Xiaodong Yang, and Jan Kautz.
\newblock {MoCoGAN: Decomposing Motion and Content for Video Generation}.
\newblock In {\em Proceedings of the IEEE Conference on Computer Vision and
  Pattern Recognition (CVPR)}, 2018.

\bibitem{ventura2020can}
Lucas Ventura, Amanda Duarte, and Xavier Gir{\'o}-i Nieto.
\newblock {Can Everybody Sign Now? Exploring Sign Language Video Generation
  from 2D Poses}.
\newblock In {\em ECCV Sign Language Recognition, Translation and Production
  Workshop}, 2020.

\bibitem{vondrick2016generating}
Carl Vondrick, Hamed Pirsiavash, and Antonio Torralba.
\newblock {Generating Videos with Scene Dynamics}.
\newblock In {\em Advances in Neural Information Processing Systems (NIPS)},
  2016.

\bibitem{wang2019example}
Miao Wang, Guo-Ye Yang, Ruilong Li, Run-Ze Liang, Song-Hai Zhang, Peter~M Hall,
  and Shi-Min Hu.
\newblock {Example-Guided Style-Consistent Image Synthesis from Semantic
  Labeling}.
\newblock In {\em Proceedings of the IEEE Conference on Computer Vision and
  Pattern Recognition (CVPR)}, 2019.

\bibitem{wang2019few}
Ting-Chun Wang, Ming-Yu Liu, Andrew Tao, Guilin Liu, Jan Kautz, and Bryan
  Catanzaro.
\newblock {Few-shot Video-to-Video Synthesis}.
\newblock In {\em Advances in Neural Information Processing Systems (NeurIPS)},
  2019.

\bibitem{wang2018video}
Ting-Chun Wang, Ming-Yu Liu, Jun-Yan Zhu, Guilin Liu, Andrew Tao, Jan Kautz,
  and Bryan Catanzaro.
\newblock {Video-to-Video Synthesis}.
\newblock In {\em Advances in Neural Information Processing Systems (NIPS)},
  2018.

\bibitem{wang2018high}
Ting-Chun Wang, Ming-Yu Liu, Jun-Yan Zhu, Andrew Tao, Jan Kautz, and Bryan
  Catanzaro.
\newblock {High-Resolution Image Synthesis and Semantic Manipulation with
  Conditional GANs}.
\newblock In {\em Proceedings of the IEEE Conference on Computer Vision and
  Pattern Recognition (CVPR)}, 2018.

\bibitem{wang2004image}
Zhou Wang, Alan~C Bovik, Hamid~R Sheikh, and Eero~P Simoncelli.
\newblock {Image Quality Assessment: From Error Visibility to Structural
  Similarity}.
\newblock {\em IEEE Transactions on Image Processing}, 2004.

\bibitem{wei2020gac}
Dongxu Wei, Xiaowei Xu, Haibin Shen, and Kejie Huang.
\newblock {GAC-GAN: A General Method for Appearance-Controllable Human Video
  Motion Transfer}.
\newblock {\em IEEE Transactions on Multimedia}, 2020.

\bibitem{wu2020mm}
Zhenyu Wu, Duc Hoang, Shih-Yao Lin, Yusheng Xie, Liangjian Chen, Yen-Yu Lin,
  Zhangyang Wang, and Wei Fan.
\newblock {MM-Hand: 3D-Aware Multi-Modal Guided Hand Generative Network for 3D
  Hand Pose Synthesis}.
\newblock {\em arXiv preprint arXiv:2010.01158}, 2020.

\bibitem{xiao2020skeleton}
Qinkun Xiao, Minying Qin, and Yuting Yin.
\newblock {Skeleton-based Chinese Sign Language Recognition and Generation for
  Bidirectional Communication between Deaf and Hearing People}.
\newblock In {\em Neural Networks}, 2020.

\bibitem{yang2018pose}
Ceyuan Yang, Zhe Wang, Xinge Zhu, Chen Huang, Jianping Shi, and Dahua Lin.
\newblock {Pose Guided Human Video Generation}.
\newblock In {\em Proceedings of the European Conference on Computer Vision
  (ECCV)}, 2018.

\bibitem{yin2020sign}
Kayo Yin.
\newblock {Sign Language Translation with Transformers}.
\newblock {\em arXiv preprint arXiv:2004.00588}, 2020.

\bibitem{zakharov2020fast}
Egor Zakharov, Aleksei Ivakhnenko, Aliaksandra Shysheya, and Victor Lempitsky.
\newblock Fast bi-layer neural synthesis of one-shot realistic head avatars.
\newblock In {\em Proceedings of the European Conference on Computer Vision
  (ECCV)}, 2020.

\bibitem{zakharov2019few}
Egor Zakharov, Aliaksandra Shysheya, Egor Burkov, and Victor Lempitsky.
\newblock {Few-Shot Adversarial Learning of Realistic Neural Talking Head
  Models}.
\newblock In {\em Proceedings of the IEEE International Conference on Computer
  Vision (CVPR)}, 2019.

\bibitem{zelinka2020neural}
Jan Zelinka and Jakub Kanis.
\newblock {Neural Sign Language Synthesis: Words Are Our Glosses}.
\newblock In {\em The IEEE Winter Conference on Applications of Computer Vision
  (WACV)}, 2020.

\bibitem{zhou2019dance}
Yipin Zhou, Zhaowen Wang, Chen Fang, Trung Bui, and Tamara Berg.
\newblock {Dance Dance Generation: Motion Transfer for Internet Videos}.
\newblock In {\em Proceedings of the IEEE International Conference on Computer
  Vision Workshops}, 2019.

\bibitem{zhu2017unpaired}
Jun-Yan Zhu, Taesung Park, Phillip Isola, and Alexei~A Efros.
\newblock {Unpaired Image-to-Image Translation using Cycle-Consistent
  Adversarial Networks}.
\newblock In {\em Proceedings of the IEEE International Conference on Computer
  Vision (ICCV)}, 2017.

\bibitem{zhu2019progressive}
Zhen Zhu, Tengteng Huang, Baoguang Shi, Miao Yu, Bofei Wang, and Xiang Bai.
\newblock {Progressive Pose Attention Transfer for Person Image Generation}.
\newblock In {\em Proceedings of the IEEE Conference on Computer Vision and
  Pattern Recognition (CVPR)}, 2019.

\end{thebibliography}
}

\end{document}